# Latent Geometry Inspired Graph Dissimilarities Enhance Affinity Propagation Community Detection in Complex Networks

Carlo Vittorio Cannistraci and Alessandro Muscoloni

*Abstract*—Affinity propagation is one of the most effective unsupervised pattern recognition algorithms for data clustering in high-dimensional feature space. However, the numerous attempts to test its performance for community detection in complex networks have been attaining results very far from the state of the art methods such as Infomap and Louvain. Yet, all these studies agreed that the crucial problem is to convert the unweighted network topology in a 'smart-enough' node dissimilarity matrix that is able to properly address the message passing procedure behind affinity propagation clustering.

Here we introduce a conceptual innovation and we discuss how to leverage network latent geometry notions in order to design dissimilarity matrices for affinity propagation community detection. Our results demonstrate that the latent geometry inspired dissimilarity measures we design bring affinity propagation to equal or outperform current state of the art methods for community detection. These findings are solidly proven considering both synthetic 'realistic' networks (with known ground-truth communities) and real networks (with community metadata), even when the data structure is corrupted by noise artificially induced by missing or spurious connectivity.

*Index Terms*—Complex networks, Community detection, Affinity propagation, Network dissimilarity, Clustering

## I. INTRODUCTION

Affinity propagation (AP) is one of the most effective algorithms for data clustering in high-dimensional feature space, however the numerous attempts to test its performance for partitioning real complex networks in communities – a task named *community detection* in network science – have been resulting until now in an 'embarrassing' collection of disillusionments. This is currently an important matter of unsupervised pattern analysis on network research that is discussed in many fields. Studies in physics, computer science and bioinformatics [1]–[5] tried to understand how to effectively apply affinity propagation for community detection but they failed, obtaining results that are far from the ones provided by the state of the art methods for community detection [6], [7] such as Infomap and Louvain. Nevertheless, all these studies agreed that the crucial problem is to convert the network topology in a 'smart-enough' node dissimilarity matrix that is able to properly address the message passing procedure behind affinity propagation clustering [1]–[3], [5]. The former solutions were based on engineering by hand or making attempts to 'try' some topological similarity/proximity measures (converted in dissimilarity/distance matrices) already known in network science, but these efforts were not guided by any network science justification to tailor the dissimilarity matrix for the message passing procedure. Here we argue that the problem should be addressed beforehand conceptually. To such aim, we recall a collection of network science notions - which are at the interface between network topology and latent network geometry [8]–[12] – based on which we propose a rationale that can guide to design dissimilarity matrices for affinity propagation community detection. Hence, we call our proposed methodology: latent geometry inspired affinity propagation (LGI-AP); and, more precisely, it deals with one of the largest studied community detection problems which, specifically, takes in consideration undirected and unweighted networks with non-overlapping communities.

Results demonstrate that the two latent geometry inspired dissimilarity measures boost affinity propagation performance to levels such that it is comparable and, on some real networks, even outperforms current state of the art methods for community detection. This is confirmed not only on the original real networks, but also when their topology is perturbed by noise simulated by random deletion of links (missing topological information) or random addition of links (spurious topological information). Finally, we perform comparative tests on artificial networks produced by the nonuniform Popularity-Similarity-Optimization (nPSO) model [13], [14]. Indeed, the nPSO is an efficient generative model recently proposed to grow realistic complex networks, which not only are clustered, small-word, scale-free and rich-club, but also present communities whose number and size can be a priory defined. These artificial networks with known community structure offer the ground-truth to build a valid benchmark to test the performance of algorithms for community detection. Even in this 'trustworthy' scenario, LPI-AP confirms to perform significantly better than the other AP versions and shows a

C. V. Cannistraci is with Biomedical Cybernetics Group, Biotechnology Center (BIOTEC), Center for Molecular and Cellular Bioengineering (CMCB), Center for Systems Biology Dresden (CSBD), Department of Physics, Technische Universität Dresden, Tatzberg 47/49, 01307 Dresden, Germany; and with Brain Bio-inspired computing (BBC) lab, IRCCS Centro Neurolesi "Bonino Pulejo", Messina, Italy (e-mail: kalokagathos.agon@gmail.com).

A. Muscoloni is with the Biomedical Cybernetics Group, Biotechnology Center (BIOTEC), Center for Molecular and Cellular Bioengineering (CMCB), Technische Universität Dresden, Tatzberg 47/49, 01307 Dresden, Germany (e-mail: alessandro.muscoloni@gmail.com).

general trend according to which its performance improves when the number of network communities grows.

## II. METHODS

### A. Affinity propagation for community detection

Affinity propagation is an algorithm for data clustering that is based on a message passing procedure. It takes as input measures of similarity (in general codified as negative dissimilarity values) between pairs of data points. Real-valued messages are exchanged between data points until a high-quality set of exemplars and corresponding clusters gradually emerges [15]. As a matter of fact, this is equivalent to exchange messages on a network where the nodes are the data points and the weights of the links are 'distances' (but often only dissimilarities) that codify a network geometry. The real problem to perform community detection by affinity propagation on a real complex network is to translate the network topology (given in a binary adjacency matrix where the value 1 indicates a link and the value 0 indicates absence of link) in a network geometry (where all the values in the adjacency matrix contain distances or most frequently only node dissimilarities). This topic will be carefully discussed in the next sections. In fact, it is not trivial to understand and define the best strategies to devise a network geometry that favours the message passing procedure exploited by affinity propagation clustering. Importantly, affinity propagation algorithm, rather than requiring the number of communities to be predefined, takes as input a real number for each network node. These values are referred to as *preferences*. Nodes with larger preferences are more likely to be chosen as exemplars of a network community. The number of identified exemplars (number of communities) is influenced by the values of the input preferences, but also emerges from the message passing procedure [15]. If a priori, all nodes are equally suitable as exemplars, the preferences should be set to a common value. This value can be varied to produce different numbers of communities. The shared preference value could be the maximum of the input similarities (resulting in a large number of communities) or their minimum (resulting in a small number of communities). Here, given in input the correct number of communities, we implemented a binary search that is able to detect the shared preference value that produces a number of communities as much as closer to the correct one. For reason of space, we invite to refer to the original article of the affinity propagation algorithm in order to check the technical details [15].

### B. Previous dissimilarity measure engineering

As aforementioned in the introduction, the former solutions were based on engineering by hand or making attempts to 'try' some topological similarity/proximity measures (converted in dissimilarity/distance matrices) already known in network science, but these efforts were not guided by any network science concept or notion to tailor the dissimilarity measure as a geometry that favours the message passing procedure of affinity propagation. Here we list four of the most employed dissimilarity measures used in the previous studies to convert the network topology in a dissimilarity matrix.

The first is the *shortest path dissimilarity (SP)* [2], [5]. It produces an adjacency (dissimilarity) matrix that collects all the shortest paths between the nodes. This measure in reality infers a network geometry which is badly approximated, because all the links in the network retain the same value one, and are not weighted with different values that suggest the node distances. Hence, it is expected a poor performance of AP related with this measure.

The second is the *Euclidean shortest path dissimilarity (ESP)* [2]. It produces an adjacency (dissimilarity) matrix by computing pairwise Euclidean distances between the nodes, where each node is characterized by a vector that collects all the shortest paths to the other nodes of the network. This measure infers a network geometry which seems more appropriate than the SP, because also the weights of the original network links will be adjusted having different real values, which approximate their pairwise distances. Hence, in theory ESP should perform better than the mere SP.

The third is the *common neighbours dissimilarity (CN)* [1], [5]. Given the original network as an adjacency matrix $A$, it produces an adjacency (dissimilarity) matrix according to the following algorithm. In the formulas below the operators ./ and .∗ indicate the element-wise division and multiplication respectively.

a. Compute the common neighbours:
$$CN = A * A$$
b. Element by element inversion to obtain a dissimilarity, only where $CN > 0$:
$$CN = 1./(1 + CN)$$
c. Compute the shortest paths of the adjacency matrix obtained at the previous point in order to get a dissimilarity matrix that contains values larger than 0 also where $CN = 0$.
d. Set the diagonal of the dissimilarity matrix to zero.

This is a typical *local* neighbourhood dissimilarity measure, because it uses topological information derived only from the first neighbours of the interacting nodes. In a previous study it gave results inferior to the next one [1]. However, here we consider it as a reference, because the CN rule is one of the most employed in network science as topological similarity.

The fourth and last measure is the *Jaccard dissimilarity (J)* [1], [5]. Given the original network as an adjacency matrix $A$, it produces an adjacency (dissimilarity) matrix according to the same algorithm of CN modified only in the first point as follow.

a. Compute the Jaccard measure:
$$J = (A * A)./U = CN./U$$

$U$ is a matrix that for each pair of nodes reports the cardinality of the union of their neighbours. In practice, it is a local neighbourhood measure equivalent to the number of common neighbours normalized for the total neighbourhood size. In a previous study [1] it gave the best results when compared versus 9 other local topological measures including CN. Here, we expect that it should confirm to perform better than CN.

### C. LPI-AP: proposed rationale and relative dissimilarity measure engineering

Our proposed rationale is that in order to favour a message passing procedure the graph dissimilarity should approximate the distances on the hidden nonlinear manifold that characterizes the graph geometry [10], [16]. The fact that the network topology emerges from this hidden geometry is the reason why many networks can efficiently send messages according to a greedy routing procedure [16]. This greedy message propagation is facilitated by the hyperbolic and tree like structure of many real complex networks [10], [16]–[18]. Recently, two pre-weighting techniques (one local and one global) based on network latent geometry have been proposed by the same authors of this study as convenient strategies for approximating the pairwise geometrical distances between connected nodes [11], [12] of an unweighted network. The trouble is that, in the previous studies, these pre-weighting techniques were devised only to confer weights to the unweighted topology of a network, without taking in consideration the problem to assign distances between disconnected nodes (dissimilarity matrix definition), which is indeed the concrete issue to address in the case of affinity propagation community detection. Here, exploiting the same latent geometry notions adopted to build the previous pre-weighting techniques, we suggest a rationale to build dissimilarity measures that contain and merge two fundamental properties that characterize the hidden geometry of many real complex networks: node network proximity (a.k.a. similarity or homophily) and node network centrality (a.k.a popularity) [10]. The node proximity is related with the node network clustering and therefore the concept of local attraction between common neighbours. The node centrality is related with the node degree and therefore with the number of neighbours that each node has. Based on this rationale, we derived from the two abovementioned pre-weighting techniques, two related (but new) dissimilarity matrices (kernels), containing dissimilarity values both for connected and disconnected nodes.

The first approach - which is called the repulsion-attraction rule (RA) [11], [12] dissimilarity – assigns an edge weight adopting only the *local* information related to its adjacent nodes (neighbourhood topological information). The idea behind RA is that adjacent nodes with a high external degree (where the external degree is computed considering the number of neighbours not in common) should be geometrically far. Indeed, they represent hubs without neighbours in common, which - according to the theory of navigability of complex networks presented by Boguñá et al. [16] - tend to dominate geometrically distant regions: this is the repulsive part of the rule. On the contrary, adjacent nodes that share a high number of common neighbours should be geometrically close because most likely they share many similarities: this is the attractive part of the rule. Thus, the RA (see below for the precise mathematical formula) is a simple and efficient approach that quantifies the trade-off between *hub repulsion* and *common-neighbours-based attraction* [11], [12]. The algorithm to compute RA dissimilarity is the following:

a. Compute the RA rule for each link in the network [11], [12] (note that this part is equivalent to the already published pre-weighting rule):

$$RA_{ij} = \frac{1 + e_i + e_j}{1 + cn_{ij}}$$

 $i$ and $j$ are the two nodes connected by a network link *i-j*, $e_i$ is the number of external links of the node *i* (links that do not connect either to common neighbours with *j* or to *j*), $e_j$ is the same for the node *j*; $cn_{ij}$ is the number of common neighbours of the link *i-j*. The numerator is the node-repulsion term, the denominator is the node-attraction term.

b. Compute the shortest paths of the adjacency matrix obtained at the previous point in order to get a dissimilarity matrix that contains values larger than 0 also for the missing links in the network (note that this part is new and introduced in this study to generate a dissimilarity matrix starting from the pre-weighting rule).

Although inspired by the same rationale, the second dissimilarity is global (exploits the entire network topology to compute each dissimilarity value between pairs of nodes), in fact as first step makes a global-information-based pre-weighting of the links, using the edge-betweenness-centrality (EBC) to approximate distances between nodes and regions of the network [11]. EBC is indeed a global topological network measure that assigns to each link a value of centrality, related to its importance in propagating information across different regions of the network. The assumption is that central edges are bridges that tend to connect geometrically distant regions of the network, while peripheral edges tend to connect nodes in the same neighbourhood. The higher the EBC value of a network link, the more information will pass through that link in message passing procedures. The algorithm to compute EBC dissimilarity is the following:

a. Compute the EBC dissimilarity for each link in the network [11]:

$$EBC_{ij} = \sum_{s,t} \frac{\sigma(s,t|e_{ij})}{\sigma(s,t)}$$

 *i* and *j* are the two nodes connected by a network link *i-j*; *s,t* is any combination of network nodes; $\sigma(s,t)$ is the number of shortest paths between *s* and *t*; $\sigma(s,t|e_{ij})$ is the number of shortest paths between *s* and *t* passing through the link $e_{ij}$. We let notice that the mere EBC pre-weighting was already adopted with very bad results [5] in a previous study that concluded this measure is not proper for AP community detection. Here, we show that transforming the manner this measure is used from a mere pre-weighting to a dissimilarity matrix is indeed an important innovation in engineering network dissimilarity measures for AP community detection. Therefore, in order to build a congruous dissimilarity measure, we introduced the next two steps.

b. Rescale the EBC pre-weighting values according to the formula:

$$EBC_{ij} = \frac{EBC_{ij}}{EBC_{ij} + \widetilde{EBC}}$$

 $\widetilde{EBC}$ represents the average EBC dissimilarity of the pre-weighted links. Differently from the other dissimilarity measures, EBC values tend to grow significantly for increasing network size, and a previous study reported problems of numerical oscillations and convergence time for this metric with AP [5], therefore a rescaling of the matrix is more appropriate. The adopted formula maps

   the values into the interval [0,1[ with the average at 0.5, while still representing a dissimilarity measure.
   c. Compute the shortest paths between nonadjacent nodes as previously reported in the step (b) for RA.

We would like to clarify that, while the pre-weighting formulas (step a) have been introduced in previous studies for improving the network embedding in the hyperbolic space [11], [12], their adoption for engineering the dissimilarity kernels to provide in input to the affinity propagation algorithm represents a novel contribution which we will show to offer significant improvements.

*D. State of the art community detection methods: Infomap and Louvain*

The community detection algorithms Infomap [19] and Louvain [20] are two state of the art approaches that have been shown to provide high performances on synthetic benchmarks [6], [21], [22]. Recently, they have been tested also on small-size and large-size real networks, resulting overall among the best performing on recovering ground-truth communities associated to metadata [29].

The Infomap algorithm [19] finds the community structure by minimizing the expected description length of a random walker trajectory using the Huffman coding process. It uses the hierarchical map equation, a further development of the map equation, to detect community structures on more than one level. The hierarchical map equation indicates the theoretical limit of how concisely a network path can be specified using a given partition structure. In order to calculate the optimal partition (community) structure, this limit can be computed for different partitions and the community annotation that gives the shortest path length is chosen. We used the C implementation released by the authors at http://www.mapequation.org/code.html.

The Louvain algorithm [20] is separated into two phases, which are repeated iteratively. At first every node in the (weighted) network represents a community in itself. In the first phase, for each node *i*, it considers its neighbours *j* and evaluates the gain in modularity that would take place by removing *i* from its community and placing it in the community of *j*. The node *i* is then placed in the community *j* for which this gain is maximum, but only if the gain is positive. If no gain is possible node *i* stays in its original community. This process is applied until no further improvement can be achieved. In the second phase the algorithm builds a new network whose nodes are the communities found in the first phase, whereas the weights of the links between the new nodes are given by the sum of the weight of the links between nodes in the corresponding two communities. Links between nodes of the same community lead to self-loops for this community in the new network. Once the new network has been built, the two phase process is iterated until there are no more changes and a maximum of modularity has been obtained. The number of iterations determines the height of the hierarchy of communities detected by the algorithm. We used the R function *multilevel.community*, an implementation of the method available in the *igraph* package [23]. For each hierarchical level there is a possible partition to compare to the ground-truth annotation. In this case, the hierarchical level considered is the one that guarantees the best match, therefore the detected partition that gives the highest NMI value. We let notice that most of this Methods section is equivalent to an analogous Methods section present in another study of the authors [11].

*E. Computational complexity*

In this section we report the computational time complexity of the community detection algorithms. The network is assumed to be connected, therefore the number of edges $E$ has at least the same order of complexity as the number of nodes $N$.

As indicated by Yang et al. [6], the complexity of the Louvain method is $O(NlogN)$, whereas the complexity of Infomap is $O(E)$.

Regarding the affinity propagation approaches, the complexity is given by two main separate steps: the construction of the dissimilarity matrix and the clustering algorithm. The dissimilarity matrix SP requires the computation of all the pairwise shortest paths between the nodes, which takes $O(EN)$ using the Johnson's algorithm [24]. The dissimilarity matrices ESP, CN, J and RA firstly require basic operations on sparse matrices, whose complexity is proportional to the number of nonzero elements, therefore $O(E)$, and secondly the computation of the shortest paths, which takes $O(EN)$. The dissimilarity matrix EBC firstly requires the computation of the edge-betweenness-centrality, which takes $O(EN)$ using the Brandes' algorithm for unweighted graphs [25], and secondly the computation of the shortest paths, which takes $O(EN)$. Summarizing, all the dissimilarity matrices have a computational complexity of $O(EN)$.

The affinity propagation algorithm, as implemented in the original publication by Frey & Dueck [15], has a cost of $O(kN^2)$, where $k$ is the number of iterations of the message passing procedure. In order to speed up the method, Jia et al. [26] proposed a fast sparse variant which reduces the complexity to $O(kN)$, at the expense of clustering accuracy. In 2011, Fujiwara et al. [27] optimized the original affinity propagation algorithm and reduced the complexity to $O(N^2+kE)$, while guaranteeing the same clustering result after convergence. However, although the complexities stated, Fujiwara et al. [27] performed simulations for computational time comparison and their proposed method resulted to be faster than the sparse variant of Jia et al. [26].

To conclude, assuming $k$ with an order of complexity at least not higher than $N$, regardless of the dissimilarity matrix adopted and of the fast affinity propagation variant implemented, the complexity of all the affinity propagation approaches is $O(EN)$.

*F. Community detection evaluation by normalized mutual information*

Different similarity measures have been developed for evaluating the matching between two partitions (the communities detected by the method and the ground-truth). They are mainly based on three categories: pair counting, cluster matching and information theory [28]. Although there is not yet one measure without any drawback, the most adopted in community detection studies is the Normalized Mutual Information (NMI) [29].

The entropy can be defined as the information contained in a distribution p(x) in the following way:

5$$H(X) = \sum_{x \in X} p(x) \log p(x)$$

The mutual information is the shared information between two distributions:

$$I(X,Y) = \sum_{y \in Y} \sum_{x \in X} p(x,y) \log \left( \frac{p(x,y)}{p_1(x) p_2(y)} \right)$$

To normalize the value between 0 and 1 the following formula can be applied:

$$NMI = \frac{I(X,Y)}{\sqrt{H(X)H(Y)}}$$

If we consider a partition of the nodes in communities as a distribution (probability of one node falling into one community), the previous equations allow us to compute the matching between the annotations obtained by the community detection algorithm and the ground-truth communities of a network. We used the MATLAB implementation available at http://commdetect.weebly.com. As suggested in the code, when $\frac{N}{n^T} \leq 100$, the NMI should be adjusted in order to correct for chance [30]. We let notice that most of this Methods section is equivalent to an analogous Methods section present in another study of the authors [11].

*G. Greedy routing*

An important measure to assess whether the distances between nodes of a network are properly estimated (in the sense that they respect the latent geometry of the network) is to test its navigability. The network is considered navigable if the greedy routing (GR) performed using the estimated node distances provides efficiency close to the one obtained when the real node distances are employed [16]. In the GR, for each pair of nodes $i$ and $j$, a packet is sent from $i$ with destination $j$. Each node knows only the address of its neighbours and the address of the destination $j$, which is written in the packet. The address of a node is represented by its coordinates in the geometrical space. At each hop the packet is forwarded from the current node to its neighbour closest to the destination. The packet is dropped when this neighbour is the same from which the packet has been received at the previous hop, since a loop has been generated. In this study, although methods for embedding networks in a geometrical space are not used, the comparison concerns dissimilarity measures, which can be considered as approximations of the distances between the nodes in a high-dimensional space. Therefore, in this particular scenario, the GR can be evaluated assessing the navigability of the network with respect to a given dissimilarity matrix: at each hop the packet is forwarded from the current node to its neighbour having the lowest dissimilarity with the destination. In order to compare the performance of the methods a GR-score has been introduced [11] and it is computed as follows:

$$GR_{score} = \frac{\sum_{i=1}^{N} \sum_{j=1, j \neq i}^{N} \frac{sp_{ij}}{p_{ij}}}{N * (N-1)}$$

Where $i$ and $j$ are two within the set of $N$ nodes, $sp_{ij}$ is the shortest path length from $i$ to $j$ and $p_{ij}$ is the GR path length from $i$ to $j$. The ratio $\frac{sp_{ij}}{p_{ij}}$ assumes values in the interval [0, 1].

We clarify that in artificial networks grown as geometrical graphs - for which is known the original geometrical location of the nodes - both $sp_{ij}$ and $p_{ij}$ are computed geometrically. It means that $sp_{ij}$ is a sum of geometrical distances in the original space and $p_{ij}$ is a path computed using the estimated distances but for which the final value is obtained as the sum of geometrical distances over the path reproduced in the original space. This is important to guarantee the scale invariance in the ratio $\frac{sp_{ij}}{p_{ij}}$. On the other hand, for real networks the real geometry is not known therefore both $sp_{ij}$ and $p_{ij}$ are computed topologically: they are obtained as a sum of steps over a path where each pair of adjacent nodes has the same distance and it fixed to one for convention. When the greedy routing is unsuccessful the path length is infinite and therefore the ratio is 0, which represents the worst case. When the greedy routing is successful the path length is greater than 0 and tends to 1 as the path length tends to the shortest path length, becoming 1 in the best case. The GR-score is the average of this ratio over all the node pairs. We let notice that most of this Methods section is equivalent to an analogous Methods section present in another study of the authors [11].

*H. Real networks datasets and statistics*

The community detection methods have been tested on 8 real networks, which represent differing systems: Karate; Opsahl_8; Opsahl_9; Opsahl_10; Opsahl_11; Polbooks; Football; Polblogs. The networks have been transformed into undirected, unweighted, without self-loops and only the largest connected component has been considered. The information of their ground-truth communities is available. Table 1, together with the results, provides also some basic statistics of the networks. $N$ is the number of nodes. $E$ is the number of edges. The parameter m refers to half of the average node degree and it is also equal to the ratio $E/N$. $C$ is the average clustering coefficient, computed for each node as the number of links between its neighbours over the number of possible links [31]. The parameter $\gamma$ is the exponent of the power-law degree distribution, fitted from the observed degree sequence using the maximum likelihood procedure developed by Clauset et al. [32] and released at http://tuvalu.santafe.edu/~aaronc/powerlaws/. The number of ground-truth communities $N_C$ is reported in Table 2.

The first network is about the Zachary's Karate Club [33], it represents the friendship between the members of a university karate club in US. The communities are formed by a split of the club into two parts, each following one trainer.

The networks from the second to the fifth are intra-organisational networks from [34] and can be downloaded at https://toreopsahl.com/datasets/#Cross_Parker. Opsahl_8 and Opsahl_9 come from a consulting company and nodes represent employees. In Opsahl_8 employees were asked to indicate how often they have turned to a co-worker for work-related information in the past, where the answers range from: 0 - I don't know that person; 1 - Never; 2 - Seldom; 3 - Sometimes; 4 - Often; 5 - Very often. Directions were ignored. The data was turned into an unweighted network by setting a link only between employees that have at least asked for information seldom (2).

*Table 1. Greedy routing on real networks.*
The table reports the greedy routing (GR) score computed for 8 real networks adopting different dissimilarity measures. The methods are ranked by mean performance over the dataset. The table contains also some statistics for each network: number of nodes $N$, number of edges $E$, clustering coefficient $C$, power law degree distribution exponent $\gamma$, half of average degree $m$. The networks are ordered for increasing size $N$ and ties are solved considering the number of edges $E$. The best result for each network as well as the best mean result are marked in bold. The SP dissimilarity measure is not reported since its GR-score is 1 by definition.

| Method | Karate $N=34$ $E=78$ $C=0.59$ $\gamma=2.12$ $m=2.29$ | Opsahl 8 $N=43$ $E=193$ $C=0.61$ $\gamma=8.20$ $m=4.49$ | Opsahl 9 $N=44$ $E=348$ $C=0.68$ $\gamma=5.92$ $m=7.91$ | Opsahl 10 $N=77$ $E=518$ $C=0.66$ $\gamma=5.06$ $m=6.73$ | Opsahl 11 $N=77$ $E=1088$ $C=0.72$ $\gamma=4.87$ $m=14.13$ | Polbooks $N=105$ $E=441$ $C=0.49$ $\gamma=2.62$ $m=4.20$ | Football $N=115$ $E=613$ $C=0.40$ $\gamma=9.09$ $m=5.33$ | Polblogs $N=1222$ $E=16714$ $C=0.36$ $\gamma=2.38$ $m=13.68$ | Mean GR-score |
|---|---|---|---|---|---|---|---|---|---|
| **EBC** | **0.99** | **1.00** | **1.00** | **0.98** | **1.00** | **0.96** | **1.00** | 0.87 | **0.97** |
| RA | 0.97 | 0.97 | 0.98 | 0.96 | 0.99 | 0.95 | 0.87 | **0.88** | 0.95 |
| ESP | 0.79 | 0.93 | 0.98 | 0.86 | 0.97 | 0.79 | 0.97 | 0.27 | 0.82 |
| J | 0.57 | 0.81 | 0.95 | 0.91 | 0.97 | 0.67 | 0.85 | 0.34 | 0.76 |
| CN | 0.56 | 0.80 | 0.91 | 0.89 | 0.97 | 0.63 | 0.81 | 0.46 | 0.75 |

In the Opsahl_9 network, the same employees were asked to indicate how valuable the information they gained from their co-worker was. They were asked to show how strongly they agree or disagree with the following statement: "In general, this person has expertise in areas that are important in the kind of work I do." The weights in this network are also based on the following scale: 0 - Do Not Know This Person; 1 - Strongly Disagree; 2 - Disagree; 3 - Neutral; 4 - Agree; 5 - Strongly Agree. We set a link if there was an agreement (4) or strong agreement (5). Directions were ignored.

The Opsahl_10 and Opsahl_11 networks come from the research team of a manufacturing company and nodes represent employees. The annotated communities indicate the company locations (Paris, Frankfurt, Warsaw and Geneva). For Opsahl_10 the researchers were asked to indicate the extent to which their co-workers provide them with information they use to accomplish their work. The answers were on the following scale: 0 – I do not know this person / I never met this person; 1 – Very infrequently; 2 – Infrequently; 3 – Somewhat frequently; 4 – Frequently; 5 – Very frequently. We set an undirected link when there was at least a weight of 4.

For Opsahl_11 the employees were asked about their awareness of each other's knowledge ("I understand this person's knowledge and skills. This does not necessarily mean that I have these skills and am knowledgeable in these domains, but I understand what skills this person has and domains they are knowledgeable in."). The weighting was on the scale: 0 – I do not know this person / I have never met this person; 1 – Strongly disagree; 2 – Disagree; 3 – Somewhat disagree; 4 – Somewhat agree; 5 – Agree; 6 – Strongly agree. We set a link when there was at least a 4, ignoring directions.

The Polbooks network represents frequent co-purchases of books concerning US politics on amazon.com. Ground-truth communities are given by the political orientation of the books as either conservative, neutral or liberal. The network is unpublished but can be downloaded at http://www-personal.umich.edu/~mejn/netdata/, as well as with the Karate, Football and Polblogs networks.

The Football network [35] presents games between division IA colleges during regular season fall 2000. Ground-truth communities are the conferences that each team belongs to.

*Table 2. Community detection on real networks.*
The table reports the Normalized Mutual Information (NMI) computed between the ground truth communities and the ones detected by every community detection algorithm for 8 real networks. NMI = 1 indicates a perfect match between the two partitions of the nodes. The methods are ranked by mean performance over the dataset. The table contains also the number of ground truth communities $N_C$. The best result for each network as well as the best mean result are marked in bold.

| Method | Karate $N_C=2$ | Opsahl 8 $N_C=7$ | Opsahl 9 $N_C=7$ | Opsahl 10 $N_C=4$ | Opsahl 11 $N_C=4$ | Polbooks $N_C=3$ | Football $N_C=12$ | Polblogs $N_C=2$ | Mean NMI |
|---|---|---|---|---|---|---|---|---|---|
| **LGI-AP-EBC** | 0.73 | 0.54 | 0.41 | 0.96 | 0.96 | 0.53 | 0.88 | **0.70** | **0.71** |
| **LGI-AP-RA** | 0.67 | 0.52 | 0.42 | **1.00** | 0.93 | **0.56** | 0.91 | 0.69 | **0.71** |
| **Infomap** | 0.55 | **0.69** | 0.47 | **1.00** | **1.00** | 0.52 | 0.92 | 0.52 | **0.71** |
| Louvain | 0.46 | 0.55 | 0.41 | **1.00** | 0.96 | 0.50 | **0.93** | 0.64 | 0.68 |
| J-AP | 0.73 | 0.48 | 0.45 | **1.00** | 0.96 | 0.39 | 0.89 | 0.40 | 0.66 |
| ESP-AP | 0.57 | 0.38 | 0.35 | 0.96 | 0.96 | 0.50 | 0.92 | 0.47 | 0.64 |
| CN-AP | 0.16 | 0.40 | **0.54** | 0.89 | 0.72 | 0.52 | 0.91 | 0.68 | 0.60 |
| SP-AP | **0.83** | 0.50 | 0.20 | 0.65 | 0.09 | 0.46 | 0.63 | 0.29 | 0.46 |

*Table 3. Community detection on real networks perturbed with random removal of links.*
For each real network, 100 perturbed networks have been generated removing at random the 10% of links. The table reports the Normalized Mutual Information (NMI) computed between the ground-truth communities and the ones detected by every community detection algorithm for the 8 real networks, averaged over the 100 repetitions. NMI = 1 indicates a perfect match between the two partitions of the nodes. The methods are ranked by mean performance over the dataset. The best result for each network as well as the best mean result are marked in bold.

| Method | Karate | Opsahl 8 | Opsahl 9 | Opsahl 10 | Opsahl 11 | Polbooks | Football | Polblogs | Mean NMI |
|---|---|---|---|---|---|---|---|---|---|
| **LGI-AP-EBC** | **0.75** | 0.51 | 0.39 | 0.98 | 0.89 | **0.54** | 0.86 | **0.70** | **0.70** |
| **LGI-AP-RA** | 0.63 | 0.51 | 0.42 | **1.00** | 0.93 | **0.54** | 0.91 | 0.69 | **0.70** |
| Infomap | 0.54 | **0.55** | **0.49** | **1.00** | **0.96** | 0.50 | **0.92** | 0.51 | 0.68 |
| Louvain | 0.49 | 0.51 | 0.42 | **1.00** | **0.96** | 0.49 | 0.90 | 0.63 | 0.68 |
| ESP-AP | 0.62 | 0.38 | 0.36 | 0.95 | 0.95 | 0.51 | 0.88 | 0.49 | 0.64 |
| J-AP | 0.34 | 0.43 | 0.40 | **1.00** | 0.93 | 0.46 | 0.89 | 0.39 | 0.60 |
| CN-AP | 0.16 | 0.37 | 0.46 | 0.91 | 0.61 | 0.53 | 0.87 | 0.66 | 0.57 |
| SP-AP | 0.63 | 0.41 | 0.21 | 0.59 | 0.15 | 0.46 | 0.67 | 0.28 | 0.42 |

The Polblogs [36] network consists of links between blogs about the politics in the 2004 US presidential election. The ground-truth communities represent the political opinions of the blogs (right/conservative and left/liberal). We let notice that most of this Methods section is equivalent to an analogous Methods section present in another study of the authors [11].

*I. Synthetic networks generated by the nPSO model*

The Popularity-Similarity-Optimization (PSO) model [10] is a generative network model recently introduced in order to describe how random geometric graphs grow in the hyperbolic space. In this model the networks evolve optimizing a trade-off between node popularity, abstracted by the radial coordinate, and similarity, represented by the angular distance. The PSO model can reproduce many structural properties of real networks: clustering, small-worldness (concurrent low characteristic path length and high clustering), node degree heterogeneity with power-law degree distribution and rich-clubness. However, being the nodes uniformly distributed over the angular coordinate, the model lacks a non-trivial community structure.

The nonuniform PSO (nPSO) model [13], [14] is a variation of the PSO model that exploits a nonuniform distribution of nodes over the angular coordinate in order to generate networks characterized by communities, with the possibility to tune their number, size and mixing property. We adopted a Gaussian mixture distribution of angular coordinates, with communities that emerge in correspondence of the different Gaussians, and the parameter setting suggested in the original study [13], [14]. Given the number of components $C$, they have means equidistantly arranged over the angular space, $\mu_i = \frac{2\pi}{C} \cdot (i - 1)$, the same standard deviation fixed to 1/6 of the distance between two adjacent means, $\sigma_i = \frac{1}{6} \cdot \frac{2\pi}{C}$, and equal mixing proportions, $\rho_i = \frac{1}{C}$ ($i = 1 \ldots C$). The community memberships are assigned considering for each node the component whose mean is the closest in the angular space. The other parameters of the model are the number of nodes $N$, half of the average node degree $m$, the network temperature $T$ (inversely related to the clustering) and the exponent $\gamma$ of the power-law degree distribution. Given the parameters ($N$, $m$, $T$, $\gamma$, $C$), for details on the generative procedure please refer to the original study [13], [14].

### III. RESULTS AND DISCUSSION

Network topologies emerging from a hidden geometry are efficiently navigable [16]. A dissimilarity matrix can be considered as an approximation of the distances between the nodes in a geometrical space and the greedy routing can be evaluated, assessing the navigability of the network under the assumption that the nodes lie in a geometrical space at the distances indicated by the dissimilarity matrix. Dissimilarity measures leading to a low navigability should not be related with a network geometry, therefore we expect a less efficient message passing procedure for affinity propagation community detection. On the other side, although a high navigability of the distance matrix should facilitate the message passing procedure, this does not necessarily imply a high performance for affinity propagation community detection. Hence, this GR-score test is intended to assess whether the proposed dissimilarity measures, as we postulated in the Methods section, are able to approximate the hidden network geometry better than the ones already available in literature. Table 1 reports the GR-score on the dataset of 8 real networks. The dissimilarity measures RA and EBC lead to the highest navigability, outperforming ESP, J and CN, in agreement with our expectations and, as we will show later, with the results in community detection. Note that the SP dissimilarity measure is not considered since its GR-score is 1 by definition. Using real networks for GR-score evaluation can be criticised because in general the ground-truth geometry (in which the network grows) is de facto unknown. To address such concern, we repeated the GR-score evaluation on realistic networks synthetically generated by the nPSO model [13], [14]. The nPSO is a random geometrical graph generative model able to produce networks with realistic structural features (such as clustering, small-worldness, power-lawness, rich-clubness) and a tailored community structure, representing a valid benchmark for many tasks such as link prediction, community detection and greedy routing. The GR-score results gathered from wide-range simulations (Fig. 1) - where synthetic networks were obtained by tuning several parameter combinations of the nPSO model – suggest that, in agreement with the results highlighted for real networks, the dissimilarity measures RA and EBC lead to the highest navigability, outperforming ESP, J and CN, and further supporting their relationship with a network geometry. In particular, RA is the most robust across several parameter combinations.

*Table 4. Community detection on real networks perturbed with random addition of links.*
For each real network, 100 perturbed networks have been generated adding at random the 10% of links. The table reports the Normalized Mutual Information (NMI) computed between the ground-truth communities and the ones detected by every community detection algorithm for the 8 real networks, averaged over the 100 repetitions. NMI = 1 indicates a perfect match between the two partitions of the nodes. The methods are ranked by mean performance over the dataset. The best result for each network as well as the best mean result are marked in bold.

| Method | Karate | Opsahl 8 | Opsahl 9 | Opsahl 10 | Opsahl 11 | Polbooks | Football | Polblogs | Mean NMI |
|---|---|---|---|---|---|---|---|---|---|
| **LGI-AP-EBC** | **0.71** | 0.48 | 0.41 | 0.97 | 0.92 | 0.49 | 0.85 | **0.56** | **0.67** |
| **LGI-AP-RA** | 0.65 | 0.51 | 0.43 | 0.98 | 0.92 | **0.55** | 0.90 | 0.45 | **0.67** |
| ESP-AP | 0.70 | 0.41 | 0.39 | 0.91 | **0.96** | 0.49 | 0.88 | 0.39 | 0.64 |
| Louvain | 0.45 | 0.51 | 0.42 | 0.98 | **0.96** | 0.49 | 0.90 | 0.41 | 0.64 |
| J-AP | 0.38 | 0.44 | 0.40 | **0.99** | 0.93 | 0.44 | 0.89 | 0.36 | 0.61 |
| CN-AP | 0.17 | 0.35 | **0.50** | 0.86 | 0.41 | 0.53 | 0.87 | 0.55 | 0.53 |
| Infomap | 0.53 | **0.55** | 0.00 | 0.98 | 0.00 | 0.50 | **0.92** | 0.31 | 0.47 |
| SP-AP | 0.59 | 0.33 | 0.18 | 0.56 | 0.13 | 0.33 | 0.61 | 0.21 | 0.37 |

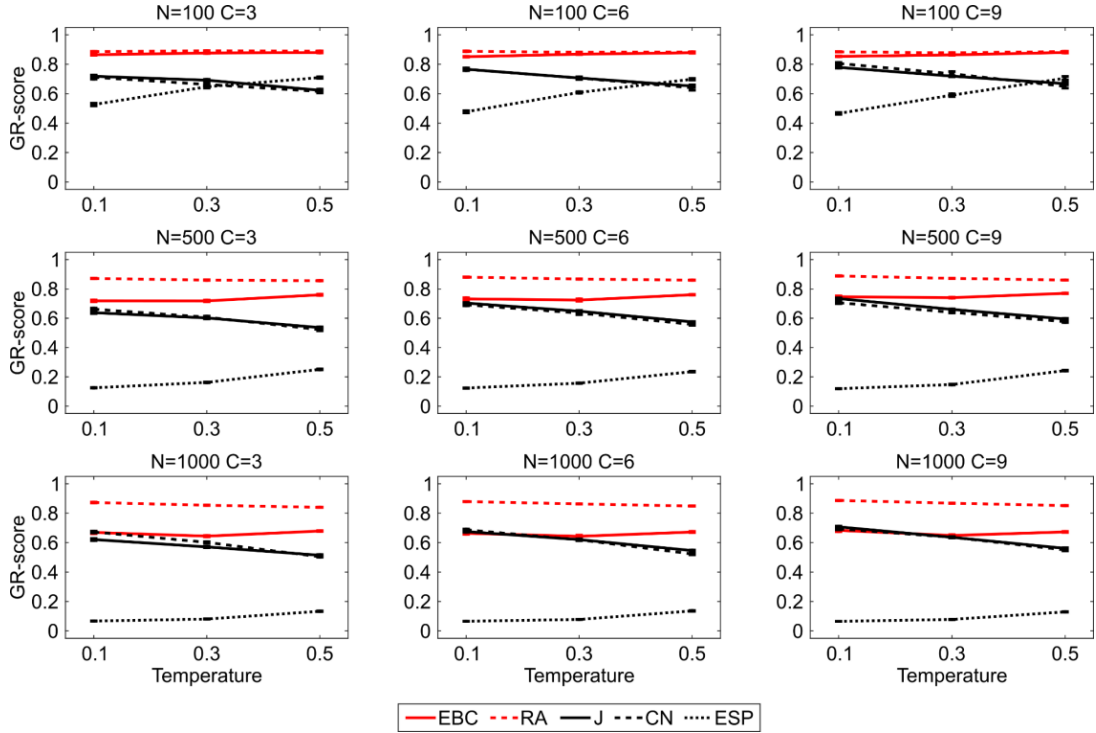

*Figure 1. Greedy routing on nPSO networks.*
Synthetic networks have been generated using the nPSO model with parameters $\gamma = 3$ (power-law degree distribution exponent), $m = 7$ (half of average degree), $T = [0.1, 0.3, 0.5]$ (temperature, inversely related to the clustering coefficient, whose respective value is reported on the upper part of each plot), $N = [100, 500, 1000]$ (network size) and $C = [3, 6, 9]$ (communities). The values of $T$, $N$ and $C$ are intended to cover the range of network characteristics observed in the dataset of small-size real networks. The value of $m$ is the average computed in the small-size real networks. Since the average $\gamma$ estimated on the dataset of small-size real networks is higher than the typical range $2 < \gamma < 3$, we choose $\gamma = 3$. For each combination of parameters, 100 networks have been generated. For each network the greedy routing has been evaluated adopting the different dissimilarity measures EBC, RA, J, CN and ESP. The plots report for each parameter combination the mean GR-score and standard error over the random repetitions. The SP dissimilarity measure is not reported since its GR-score is 1 by definition.

In Table 2 we report the comparison of the four previously published dissimilarity measures, the two new proposed dissimilarity measures (RA and EBC) and the state of the art methods for community detection Infomap and Louvain. The 8 considered networks represent a benchmark with ground-truth annotation up today available and generally adopted to test algorithm for non-overlapping community detection on real network topologies. However, the results we obtain suggest that this benchmark, collecting networks of different size (from tenths to thousands of nodes), seems enough complete and diversified to adequately investigate the performance of each dissimilarity matrix. In fact, according to the rationale described in the Methods, the SP dissimilarity should be the method with the worst performance, because does not suggest a 'congruous' network geometry that supports the message passing procedure. The ESP, instead, should offer better results than pure SP, because approximates a network geometry. This theoretical expectations are confirmed by the results in Table 1. Similarly, CN should perform worse than J dissimilarity, because J conveys in the numerator an attractive rule (it is CN) and in the denominator a repulsive rule (function of the nodes' degree), that according to our rationale should favour the inference of a network geometry that supports the message passing procedure. In fact, in our tests, J largely outperform CN and offers approximatively the same performance of ESP. Remarkably, both RA and EBC display high performance, comparable to the state of the art method Infomap and higher than Louvain, confirming the validity of our rationale on how to design dissimilarity measures that favour the message passing procedure of affinity propagation for community detection.

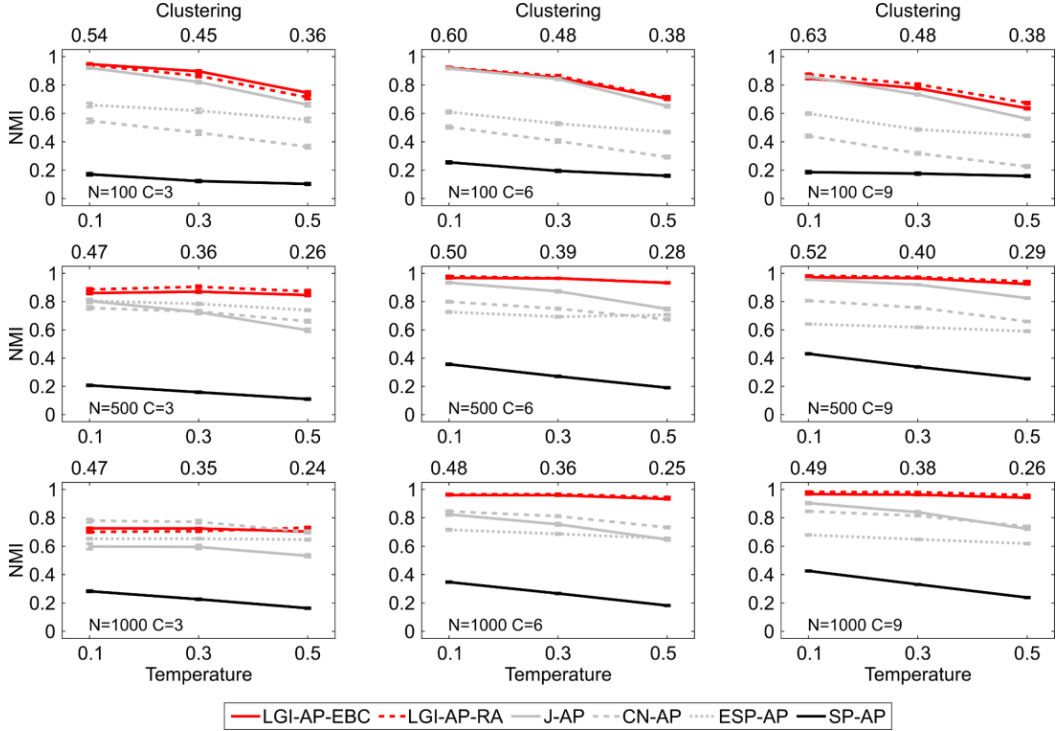

*Figure 2. Community detection on nPSO networks: comparison between different affinity propagation variants.*
Synthetic networks have been generated using the nPSO model with parameters $\gamma = 3$ (power-law degree distribution exponent), $m = 7$ (half of average degree), $T = [0.1, 0.3, 0.5]$ (temperature, inversely related to the clustering coefficient, whose respective value is reported on the upper part of each plot), $N = [100, 500, 1000]$ (network size) and $C = [3, 6, 9]$ (communities). The values of $T$, $N$ and $C$ are intended to cover the range of network characteristics observed in the dataset of small-size real networks. The value of $m$ is the average computed in the small-size real networks. Since the average $\gamma$ estimated on the dataset of small-size real networks is higher than the typical range $2 < \gamma < 3$, we choose $\gamma = 3$. For each combination of parameters, 100 networks have been generated. For each network the community detection methods LGI-AP-RA, LGI-AP-EBC, J-AP, CN-AP, ESP-AP and SP-AP have been executed and the communities detected have been compared to the annotated ones computing the Normalized Mutual Information (NMI). The plots report for each parameter combination the mean NMI and standard error over the random repetitions.

Interestingly, we made other two in-silico experiments on community detection to test the robustness of our findings also in case of noise injection in the real networks topology. In the first case we perturbed the real network structure by random deletion of 10% of the original number of network links. We repeated this procedure for 100 realizations, and the average results are reported in Table 3. This experiment simulates the behaviour of our algorithms in case of partial (10%) missing topological information. The results suggest that most of the methods reduce their performance, but both our methods outperform Infomap and Louvain. In the second case we perturbed the real network structure by random addition of 10% of the original number of network links. We repeated this procedure for 100 realizations, and the average results are reported in Table 4. This experiment simulates the behaviour of our algorithms in case of partial (10%) addition of wrong topological information. The results suggest that almost all the methods reduce again their performance, but here the decrease is more significant than in the case of missing information, pointing out that the injection of wrong links can remarkably impair community detection. In particular, we notice that Infomap is the algorithm that suffers more the injection of false links, indeed in Opsahl_9 and Opsahl_11 it detects the whole network as a unique community (NMI = 0).

Finally, in order to provide further convincing results, we tested the community detection methods considering again realistic networks synthetically generated by the nPSO model [13], [14] (as we did at the beginning for the greedy routing evaluation), because they present a tailored community structure, which is a valid benchmark for ground-truth-based community detection evaluation. The results of wide-range simulations (Figs. 2 and 3) - where synthetic networks were obtained by tuning several parameter combinations of the nPSO model - highlight also here similarities with respect to the results on real networks. First, LGI-AP-EBC and LGI-AP-RA overall offer similar performances, and both of them outperform in general the other AP versions that are not latent geometry inspired (Fig. 2). Second, both LGI-AP versions perform better than the state of the art methods (Fig. 3) for some parameter combinations (in particular for bigger network size $N = 1000$, more communities $C = 6\text{-}9$, and low temperature $T = 0.1$), whereas for other parameter combinations they are either equivalent, or Louvain and Infomap perform slightly better. Hence, the methods have overall comparable performances, which implies that the results obtained on real networks are well endorsed also by the experiments on synthetic networks.

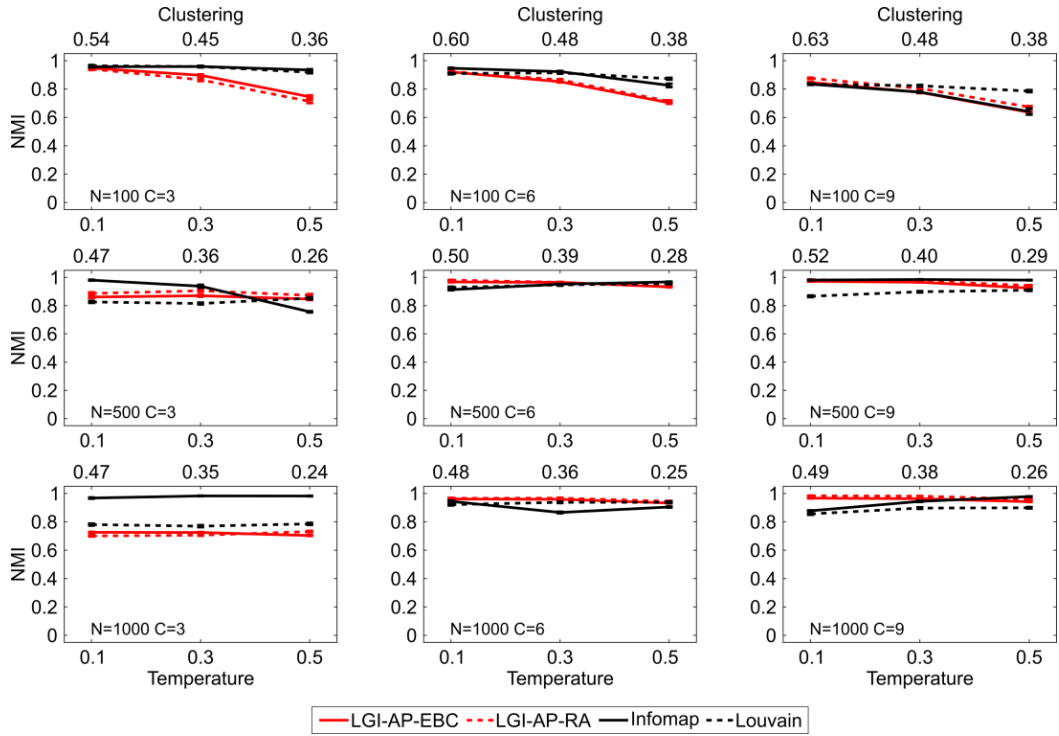

*Figure 3. Community detection on nPSO networks: comparison between LGI-AP and state of the art algorithms.*
Synthetic networks have been generated using the nPSO model with parameters $\gamma = 3$ (power-law degree distribution exponent), $m = 7$ (half of average degree), $T = [0.1, 0.3, 0.5]$ (temperature, inversely related to the clustering coefficient, whose respective value is reported on the upper part of each plot), $N = [100, 500, 1000]$ (network size) and $C = [3, 6, 9]$ (communities). The values of $T$, $N$ and $C$ are intended to cover the range of network characteristics observed in the dataset of small-size real networks. The value of $m$ is the average computed in the small-size real networks. Since the average $\gamma$ estimated on the dataset of small-size real networks is higher than the typical range $2 < \gamma < 3$, we choose $\gamma = 3$. For each combination of parameters, 100 networks have been generated. For each network the community detection methods LGI-AP-RA, LGI-AP-EBC, Louvain and Infomap have been executed and the communities detected have been compared to the annotated ones computing the Normalized Mutual Information (NMI). The plots report for each parameter combination the mean NMI and standard error over the random repetitions.

In conclusion, this article proposes a rationale on how to design dissimilarity measures, which starting from the pure topology tries to approximate the hidden geometry of the manifold that generates the network topology. Since the hidden geometry of many real complex networks is hyperbolic and tree-like [10], [16], its congruous approximation can favour the message passing procedure of algorithms such as affinity propagation for community detection in real complex networks. The empirical and numerical results provided in this study support the proposed rationale, and the two derived dissimilarity measures RA and EBC seem to boost affinity propagation to a level that is comparable with the current state of the art. Future studies could focus to evolve the present LGI-AP method, which suffers issues such as the time complexity (which is around $O(EN)$) and the number of iterations to converge to the final result, which represent concrete problems on large size networks (number of nodes much larger than 1000). Altogether, our study suggests that LGI-AP, if adequately developed, is a good candidate to become a new state of the art algorithm for community detection in real complex networks.

HARDWARE AND SOFTWARE

MATLAB code has been used for all the simulations, except for the Infomap community detection (C implementation). MATLAB simulations have been carried out in a workstation under Windows 8.1 Pro with 512 GB of RAM and 2 Intel(R) Xenon(R) CPU E5-2687W v3 processors with 3.10 GHz. C simulations have been carried out in the ZIH-Cluster Taurus of the TU Dresden.

ACKNOWLEDGMENT

We thank the BIOTEC System Administrators for their IT support and the Centre for Information Services and High Performance Computing (ZIH) of the TUD. We thank Gloria Marchesi for the administrative assistance.

FUNDING

The work was supported by the independent research group leader starting grant of the Technische Universität Dresden. AM was partially supported by the funding provided by the Free State of Saxony in accordance with the Saxon Scholarship Program Regulation, awarded by the Studentenwerk Dresden based on the recommendation of the board of the Graduate Academy of TU Dresden.

AUTHOR CONTRIBUTIONS

CVC invented the latent geometry inspired graph dissimilarities and designed the numerical experiments. CVC implemented the main function for community detection and performed the analysis on real networks. AM performed the analysis on the perturbed real networks and on synthetic networks. Both the authors analyzed and interpreted the results. CVC wrote the main sections of the article and AM integrated with the secondary sections. CVC designed the figures and tables and AM realized them. CVC planned, directed and supervised the study.

COMPETING INTERESTS

The authors declare no competing financial interests.

REFERENCES

[1] W. F. Guo and S. W. Zhang, "A general method of community detection by identifying community centers with affinity propagation," *Phys. A Stat. Mech. its Appl.*, vol. 447, pp. 508–519, 2016.

[2] H. W. Liu, "Community detection by affinity propagation with various similarity measures," *Proc. - 4th Int. Jt. Conf. Comput. Sci. Optim. CSO 2011*, pp. 182–186, 2011.

[3] Y. Shuzhong and L. Siwei, "Community detection based on adaptive kernel affinity propagation," *Comput. Sci. Inf. Technol. 2009. ICCSIT 2009. 2nd IEEE Int. Conf.*, vol. 22, no. 2013, pp. 1–4, 2009.

[4] J. Vlasblom and S. J. Wodak, "Markov clustering versus affinity propagation for the partitioning of protein interaction graphs.," *BMC Bioinformatics*, vol. 10, p. 99, 2009.

[5] Z. Liu, P. Li, Y. Zheng, and M. Sun, "Community Detection by Affinity Propagation," *Work*, p. 12, 2008.

[6] Z. Yang, R. Algesheimer, and C. J. Tessone, "A Comparative Analysis of Community Detection Algorithms on Artificial Networks," *Sci. Rep.*, vol. 6, p. 30750, 2016.

[7] D. Hric, R. K. Darst, and S. Fortunato, "Community detection in networks: Structural communities versus ground truth," *Phys. Rev. E - Stat. Nonlinear, Soft Matter Phys.*, vol. 90, no. 6, 2014.

[8] M. Á. Serrano, D. Krioukov, and M. Boguñá, "Self-similarity of complex networks and hidden metric spaces," *Phys. Rev. Lett.*, vol. 100, no. 7, pp. 1–4, 2008.

[9] D. Krioukov, F. Papadopoulos, M. Kitsak, A. Vahdat, and M. Boguñá, "Hyperbolic geometry of complex networks," *Phys. Rev. E - Stat. Nonlinear, Soft Matter Phys.*, vol. 82, no. 3, p. 036106, 2010.

[10] F. Papadopoulos, M. Kitsak, M. A. Serrano, M. Boguñá, and D. Krioukov, "Popularity versus similarity in growing networks," *Nature*, vol. 489, no. 7417, pp. 537–540, 2012.

[11] A. Muscoloni, J. M. Thomas, S. Ciucci, G. Bianconi, and C. V. Cannistraci, "Machine learning meets complex networks via coalescent embedding in the hyperbolic space," *Nat. Commun.*, vol. 8, 2017.

[12] A. Muscoloni and C. V. Cannistraci, "Minimum curvilinear automata with similarity attachment for network embedding and link prediction in the hyperbolic space," *arXiv:1802.01183 [physics.soc-ph]*, 2018.

[13] A. Muscoloni and C. V. Cannistraci, "A nonuniform popularity-similarity optimization (nPSO) model to efficiently generate realistic complex networks with communities," *New J. Phys.*, vol. 20, 2018.

[14] A. Muscoloni and C. V. Cannistraci, "Leveraging the nonuniform PSO network model as a benchmark for performance evaluation in community detection and link prediction," *New J. Phys.*, vol. 20, 2018.

[15] B. J. Frey and D. Dueck, "Clustering by Passing Messages Between Data Points," *Science (80-. ).*, vol. 315, no. 5814, pp. 972–976, 2007.

[16] M. Boguñá, D. Krioukov, and K. C. Claffy, "Navigability of complex networks," *Nat. Phys.*, vol. 5, no. 1, pp. 74–80, 2008.

[17] C. V. Cannistraci, G. Alanis-Lobato, and T. Ravasi, "Minimum curvilinearity to enhance topological prediction of protein interactions by network embedding," *Bioinformatics*, vol. 29, no. 13, pp. 199–209, 2013.

[18] C. V. Cannistraci, T. Ravasi, F. M. Montevecchi, T. Ideker, and M. Alessio, "Nonlinear dimension reduction and clustering by Minimum Curvilinearity unfold neuropathic pain and tissue embryological classes," *Bioinformatics*, vol. 26, pp. i531–i539, 2010.

[19] M. Rosvall and C. T. Bergstrom, "Multilevel compression of random walks on networks reveals hierarchical organization in large integrated systems," *PLoS One*, vol. 6, no. 4, p. e18209, 2011.

[20] V. D. Blondel, J.-L. Guillaume, R. Lambiotte, and E. Lefebvre, "Fast unfolding of communities in large networks," *J. Stat. Mech Theory Exp.*, vol. 2008, no. 10, p. 10008, 2008.

[21] G. K. Orman and V. Labatut, "A Comparison of Community Detection Algorithms on Artificial Networks," in *Discovery science*, 2009, pp. 242–256.

[22] A. Lancichinetti and S. Fortunato, "Community detection algorithms: A comparative analysis," *Phys. Rev. E*, vol. 80, no. 5, p. 056117, 2009.

[23] G. Csárdi and T. Nepusz, "The igraph software package for complex network research," *InterJournal Complex Syst.*, vol. 1695, 2006.

[24] D. B. Johnson, "Efficient Algorithms for Shortest Paths in Sparse Networks," *J. ACM*, vol. 24, no. 1, pp. 1–13, 1977.

[25] U. Brandes, "A faster algorithm for betweenness centrality," *J. Math. Sociol.*, vol. 25, no. 2, pp. 163–177, 2001.

[26] Y. Jia, J. Wang, C. Zhang, and X.-S. Hua, "Finding image exemplars using fast sparse affinity propagation," in *Proceeding of the 16th ACM international conference on Multimedia - MM '08*, 2008, p. 639.
[27] Y. Fujiwara, G. Irie, and T. Kitahara, "Fast Algorithm for Affinity Propagation," *Ijcai*, pp. 2238–2243, 2011.
[28] S. Fortunato and D. Hric, "Community detection in networks: A user guide," *Phys. Rep.*, vol. 659, pp. 1–44, 2016.
[29] L. Danon, A. Diaz-Guilera, J. Duch, and A. Arenas, "Comparing community structure identification," *J. Stat. Mech. Theory Exp.*, vol. P09008, pp. 1–10, 2005.
[30] N. X. Vinh, J. Epps, and J. Bailey, "Information Theoretic Measures for Clusterings Comparison: Variants, Properties, Normalization and Correction for Chance," *J. Mach. Learn. Res.*, vol. 11, pp. 2837–2854, Dec. 2010.
[31] D. J. Watts and S. H. Strogatz, "Collective dynamics of 'small-world' networks," *Nature*, vol. 393, no. 6684, pp. 440–442, 1998.
[32] A. Clauset, C. Rohilla Shalizi, and M. E. J. Newman, "Power-Law Distributions in Empirical Data," *SIAM Rev.*, vol. 51, no. 4, pp. 661–703, 2009.
[33] W. W. Zachary, "An Information Flow Model for Conflict and Fission in Small Groups," *J. Anthropol. Res.*, vol. 33, no. 4, pp. 452–473, 1977.
[34] R. Cross and A. Parker, *The Hidden Power of Social Networks*, no. October. 2004.
[35] M. Girvan and M. E. J. Newman, "Community structure in social and biological networks," *PNAS*, vol. 99, no. 12, pp. 7821–7826, 2002.
[36] L. A. Adamic and N. Glance, "The Political Blogosphere and the 2004 U.S. Election: Divided They Blog," *LinkKDD 2005*, pp. 36–43, 2005.

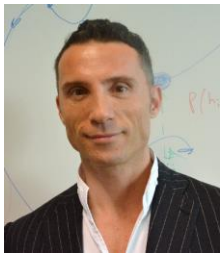
**Carlo Vittorio Cannistraci** is a theoretical engineer and was born in Milazzo, Sicily, Italy in 1976. He received the M.S. degree in Biomedical Engineering from the Polytechnic of Milano, Italy, in 2005 and the Ph.D. degree in Biomedical Engineering from the Inter-polytechnic School of Doctorate, Italy, in 2010.

From 2009 to 2010, he was visiting scholar in the Integrative Systems Biology lab of Dr. Trey Ideker at the University California San Diego (UCSD), CA, USA. From 2010 to 2013, he was Post-doc and then Research Scientist in machine intelligence and complex network science for personalized biomedicine at the King Abdullah University of Science and Technology (KAUST), Saudi Arabia. Since 2014, he has been Independent Group Leader and Head of the Biomedical Cybernetics lab at the Biotechnological Center (BIOTEC) of the TU-Dresden, Germany. He is author of three book chapters and more than 40 articles. His research interests include subjects at the interface between physics of complex systems, complex networks and machine learning theory, with particular interest for applications in biomedicine and neuroscience.

Dr. Cannistraci is member of the Network Science Society, member of the International Society in Computational Biology, member of the American Heart Association, member of the Functional Annotation of the Mammalian Genome Consortium. He is an Editor for the mathematical physics board of the journal Scientific Reports edited by Nature. *Nature Biotechnology* selected his article (*Cell* 2010) on machine learning in developmental biology to be nominated in the list of 2010 notable breakthroughs in computational biology. *Circulation Research* featured his work (*Circulation Research* 2012) on leveraging a cardiovascular systems biology strategy to predict future outcomes in heart attacks, commenting: "a space-aged evaluation using computational biology". In 2017, Springer-Nature scientific blog highlighted with an interview to Dr. Cannistraci his recent study on "How the brain handles pain through the lens of network science". In 2018, the American Heart Association covered on its website Dr. Cannistraci's chronobiology discovery on how the sunshine affects the risk and time onset of heart attack. The TU-Dresden honoured Dr. Cannistraci of the *Young Investigator Award 2016 in Physics* for his recent work on the local-community-paradigm theory and link prediction in monopartite and bipartite networks.

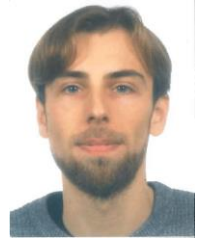
**Alessandro Muscoloni** was born in Italy in 1992. He received the B.S. degree in Computer Engineering (2014) and the M.S. degree in Bioinformatics (2016) from the University of Bologna, Italy.

He is currently a computer science Ph.D. student under the supervision of Dr. Carlo Vittorio Cannistraci in the Biomedical Cybernetics Lab at the Biotechnology Center (BIOTEC) of the TU-Dresden, Germany. He recently published in 2017 as first author together with Dr. Cannistraci as last and corresponding author an article on an efficient machine learning algorithm for the embedding of complex networks in the hyperbolic space: "Machine learning meets complex networks via coalescent embedding in the hyperbolic space", Nature Communications 8, 1615 (2017), doi:10.1038/s41467-017-01825-5. His research interests include complex networks and machine learning.

Mr. Muscoloni was awarded a scholarship to meritorious students enrolled at the University of Bologna in the AY 2015/2016, and he was selected by the Graduate Academy of TU-Dresden as awardee of the Saxon Scholarship Program (Sept. 2016 – Aug. 2017) and twice of the Travel Award for Conferences (Dec. 2016, Dec. 2017).

Mr. Muscoloni is member of the Network Science Society.